\documentclass[sigconf,authorversion,nonacm]{acmart}

\AtBeginDocument{%
  }

\usepackage{xspace}
\usepackage[symbol]{footmisc}
\usepackage{amsmath}
\newcommand{\system}{\textsc{$\text{PrivacyBrain}$}\xspace}
\newcommand{\pb}{\textsc{$\text{PB}$}\xspace}
\newcommand{\point}[1]{\vspace{.05in} \par\noindent\textbf{#1}. }

\begin{document}

\title{Ingest-And-Ground: Dispelling Hallucinations from Continually-Pretrained LLMs with RAG}

\author{Chenhao Fang*, Derek Larson*, Shitong Zhu*, Sophie Zeng*, Wendy Summer*, Yanqing Peng*, Yuriy Hulovatyy*, Rajeev Rao, Gabriel Forgues, Arya Pudota, Alex Goncalves, Hervé Robert}
\email{{chenhaofang, dereklarson, shitong, sophiezeng, wsummer, yanqingpeng, jura, rrrao, gforgues, arpu, alexgon, hervert}@meta.com}
\affiliation{%
  \institution{Meta}
  \city{Menlo Park}
  \state{California}
  \country{USA}
}

\renewcommand{\shortauthors}{Fang et al.}

\begin{abstract}
This paper presents new methods that have the potential to improve privacy process efficiency
with LLM and RAG. To reduce hallucination, we continually pre-train the base LLM model with a privacy-specific knowledge base and then augment it with a semantic RAG layer. Our evaluations demonstrate that this approach enhances the model performance (as much as doubled metrics compared to out-of-box LLM) in handling privacy-related queries, by grounding responses with factual information which reduces inaccuracies.
\end{abstract}

\maketitle
\footnotetext[1]{Authors contributed equally}

\section{Introduction}

Robust privacy compliance~\cite{metaPrivacy} is achieved through a collaborative privacy review process for projects involving engineers and privacy regulation experts~\cite{metaPrivacyProgress}. Although this method is effective, it is also time-consuming and labor-intensive. More specifically, engineers may possess a diverse range of knowledge regarding these privacy regulations, leading to multiple project revisions to ensure compliance. Meanwhile, reviewers must also grasp the technical details, resulting in numerous queries to thoroughly comprehend the projects.

\point{Goals} Our mission is to develop tools that assist both parties in this process. However, privacy regulations are too complex for traditional rule-based or AI techniques, since they require interpretation across human language, code, and both structured and unstructured data. 
The advent of Large Language Models (LLMs) offers a promising solution, since they are adept at understanding different types of inputs~\cite{nam2024using}, can perform logical reasoning and handle complex tasks~\cite{suzgun2022challenging}, and are interactive which allow them to provide explanations for their assessments. 
An LLM equipped with internal privacy knowledge could help engineers and privacy moderators find answers to privacy questions more efficiently, greatly streamlining privacy reviews. 

\point{Challenges} 
The concept of "privacy risk" is dynamic and
even the most advanced model can struggle with the specialized and ever-changing nature of privacy regulations and enterprise-specific data structures, leading to errors and irrelevant responses~\cite{achiam2023gpt}. This phenomenon, known as "hallucination"~\cite{huang2023surveyhallucinationlargelanguage}, can significantly undermine the usability of the model. For example, in assessing data storage regulations, a LLM model experiencing hallucinations might erroneously assert compliance based on non-existent regulations or a misinterpreted data storage component. Unlike other common issues in LLM responses such as logical errors and knowledge gaps, hallucinations are particularly problematic in our context as they appear correct at first glance and are tricky for engineers and privacy reviewers to immediately recognize inaccuracies in the responses. This results in wasted time and resources as both parties need to verify the correctness of generated response.


\point{Solution} To address these challenges, Retrieval-Augmented Generation (RAG) offers a promising solution~\cite{shuster2021retrievalaugmentationreduceshallucination}. RAG systems integrate information retrieval with text generation, enabling them to dynamically access the most current knowledge bases and retrieve relevant context data in real-time~\cite{lewis2020retrieval}. This approach ensures that the responses generated are not only contextually appropriate but also reflect the latest standards, while providing citations for their conclusions. 
In practice, effective RAG systems 
have been proven to significantly reduce the risk of hallucinations in different use cases~\cite{shuster2021retrieval}, and we explore a similar solution to build a system called \system to improve the LLM-generated answers for privacy compliance assessment.

\section{Methodology}
For this reserach, we choose to continually pre-train \texttt{Llama-3.1}~\cite{dubey2024llama} to ingest the privacy knowledge base 
for enhancing the base pre-trained model on fronts of understanding and reasoning for various privacy-related tasks (e.g., identifying regulatory risks for a new product launch). While the original Llama3.1 family showed their state-of-the-art performance, including factuality assessments which measures how vulnerable the models are prone to hallucination issues~\cite{dubey2024llama}, the additional round of pre-training can elevate them again, as the new training updates model weights without calibrating the model through any alignment process (e.g., RLHF~\cite{ouyang2022training}). Given so, we aim to ground hallucinations with a proper RAG system that provides factual information to the LLM, by retrieving it from a knowledge base.

\vspace{-0.1cm}

\begin{figure}[ht]
    \centering
    \includegraphics[width=0.5\textwidth]{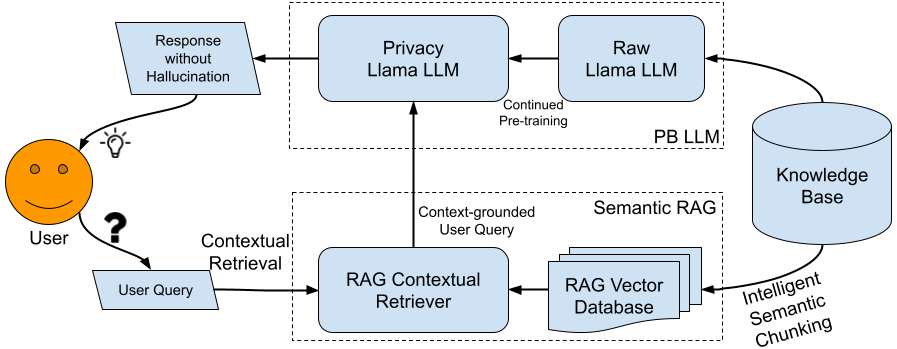} 
    \caption{System overview of \system: how we factually ground LLM hallucinations with RAG}
    \label{fig:SysOvr}
    \vspace{-2mm}
\end{figure}

As depicted in Figure~\ref{fig:SysOvr}, \system includes the following main components/steps:
\point{Privacy knowledge base} First, we need to compose the dataset that covers our desirable privacy-related documentations, to serve as the data source for both performing additional model training and providing document pieces to RAG. We follow guidance from privacy experts to construct a knowledge base of about 20,000 documents (which translates to roughly 2 million tokens) that encapsulates privacy policies and processes, as well as privacy laws and regulations for various countries and regions. 

\point{LLM continual pre-training} We use \texttt{Llama3.1-70b-instruct} as the base model, and perform continual pre-training over it using the collected documents in the knowledge base. We follow the general pre-training paradigm reported in~\cite{dubey2024llama}, which utilizes Causal Token Masking as the training task. 

\point{RAG with semantic chunking} To reduce hallucinations, we further use the constructed knowledge base to build an intelligent RAG layer on top of the enhanced LLM. When presented with a query prompt, the RAG layer retrieves relevant documents or passages from the knowledge base and then uses both the original prompt and the retrieved information to generate a coherent and contextually enriched response. One key step of indexing documents for RAG systems is how to perform chunking across long documents, as it directly impacts the effectiveness of retrievers to locate the most relevant document pieces. Inappropriate segmentation strategies can result in chunks that either lack sufficient context or contain too much irrelevant information, thereby impairing the performance of retrieval models~\cite{shi2023largelanguagemodelseasily}.

Traditional chunking methods, such as segmenting by sentences or paragraphs~\cite{characterSplitting}, typically generate snippets of uniform or similar sizes. However, they often fall short in considering text semantics, leading to sub-optimal retrieval performance. More sophisticated techniques include recursive character splitting~\cite{recursiveSplitting}, which segments texts based on a hierarchy of delimiters like paragraph breaks and spaces, respect documents' intrinsic structure better, but still compromises contextual richness. More recent algorithms are semantic-based splitting (e,g.,~\cite{semanticChunking}), which uses text embeddings to group text segments with similar meanings. These approaches segments documents at ``semantically-natural'' breaks such as sentence endings, ensuring that each chunk is contextually coherent and semantically connected. In this work, we implemented our semantic RAG module based on Meta's state-of-the-art text embedding model, Dragon-Plus~\cite{lin2023train}.

\point{Online context grounding} Now, with both the enhanced LLM and semantic RAG module ready, we need to put them into action in the online environment. Specifically, we feed example queries related to privacy first to the RAG, which retrieves fact-grounded and relevant document chunks as additional context to enhance the original queries, and then pass them to LLM. With the additional context that grounds the factuality of queries provided, LLM now stands much lower odds to hallucinate, as will be shown in evaluations.



\section{Evaluations}


\point{Data}
We construct a privacy understanding benchmark that consists of 50 example queries (e.g., ``What are principles for protecting user data in general, for a social network company?''), which covers a wide range of common questions about privacy(e.g., privacy processes, privacy laws/regulations). We manually construct ground truth answer text for each question according to references, as well as a short list of keywords that capture the key points from the answer text. 

\point{Quality measurement \& metric}
Assessing quality of LLM systems, especially based on Q\&A interactions, has been gaining increasing popularity, as it is intrinsically challenging. We tackle this by combining two popular and widely-adopted mechanisms: (i) \textit{LLM-as-a-Judge rating}~\cite{zheng2023judgingllmasajudgemtbenchchatbot}, which invokes a potent third-party LLM (GPT-4 in our case) to rate each answer generated by \system, with respect to the compiled ground truth. The metric for this measurement is \textit{pass rate}. (ii) \textit{Keyword matching}, which relies on the manually-extracted short list of key points to determine how many of them are present in the LLM response. We measure this metric by $\frac{|\textit{matched keywords}|}{|\textit{total keywords}|}$. We believe this combination offers a reasonable approximation to high-quality human inspection, which is otherwise costly to carry out. 

\point{Baselines}
We choose three baselines to evaluate the performance \system (or \pb in short) in relation to its variants, to showcase improvements by incorporating different components: (i) raw Llama3.1 with continual pre-training (\pb LLM in short); (ii) raw Llama3.1 with semantic RAG; and (iii) raw Llama3.1.

\vspace{-0.3cm}




\begin{figure}[ht]
    \centering
    \includegraphics[width=0.5\textwidth]{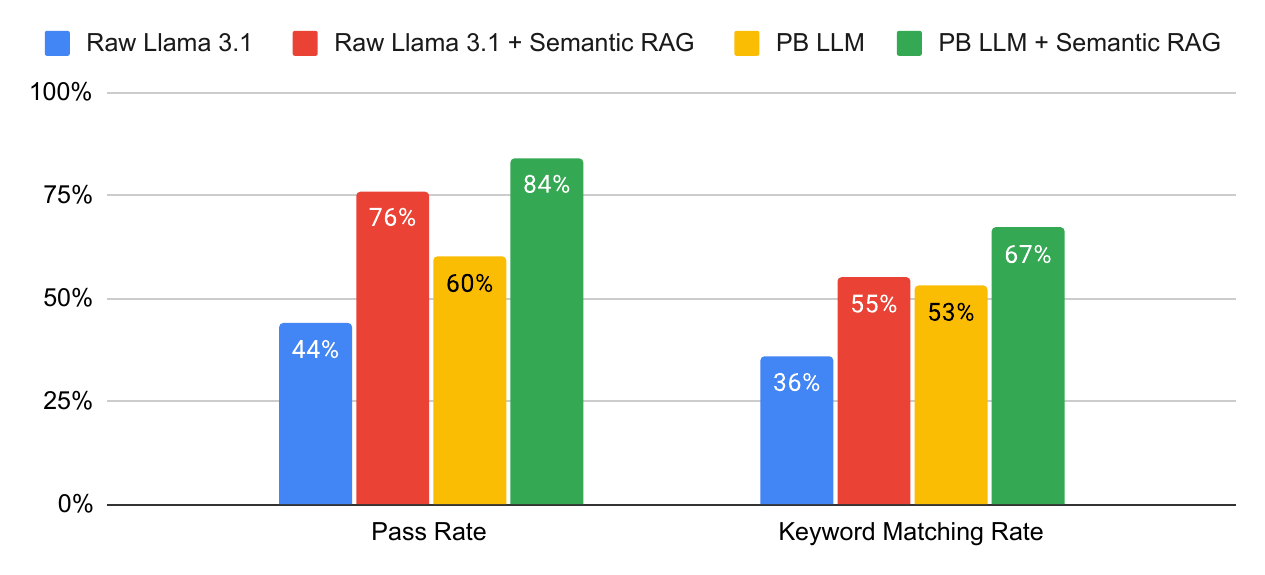} 
    \vspace{-1cm}
    \caption{Performance comparison with baselines}
    \label{fig:results}
\end{figure}
\vspace{-0.5cm}


\point{Results}
Figure~\ref{fig:results} highlights key observations based on our evaluation set: (i) continual pre-training significantly (+16\%) helps Llama LLM 
ingest domain knowledge that it was not initially trained with, resulting in higher rates passing both GPT4 and keyword-based quality measurements; (ii) semantic RAG visibly tames hallucinations that cause factual errors generated from the \system LLM, which translates to further improvements (+24\% vs. PB LLM only, and 40\% vs. raw Llama) over metrics in measuring the correctness of LLM responses.
Collectively, we conclude that the combination of the knowledge injection via additional LLM pre-training, and the contextual RAG layer (note the additional knowledge powering these two enhancements is from the same source) strengthens the overall performance of \system.

\section{Conclusion}
Our preliminary results demonstrate the potential of LLM models with RAG systems in enhancing privacy compliance in enterprise settings. 
Future work will focus on expanding the data sources, improving the retrieval mechanisms, and further fine-tuning the system for specific sub-domains within privacy laws/regulations.

\bibliography{acmart}

\end{document}